\documentclass[sigconf]{acmart}

\AtBeginDocument{%
  \providecommand\BibTeX{{%
    \normalfont B\kern-0.5em{\scshape i\kern-0.25em b}\kern-0.8em\TeX}}}

\setcopyright{acmcopyright}
\copyrightyear{2022}
\acmYear{2022}
\acmDOI{XXXXXXX.XXXXXXX}

\acmConference[FIRE '22]{Forum for Information Retrieval and Evaluation}{December 09-12, 2022}{Delhi, India}
\copyrightyear{2022}
\acmYear{2022}
\setcopyright{acmcopyright}\acmConference[FIRE '22]{Forum for Information
Retrieval Evaluation}{December 9--13, 2022}{Kolkata, India}
\acmBooktitle{Forum for Information Retrieval Evaluation (FIRE '22), December
9--13, 2022, Kolkata, India}
\acmPrice{15.00}
\acmDOI{10.1145/3574318.3574334}
\acmISBN{979-8-4007-0023-1/22/12}
\begin{document}

\title{Causal Categorization of Mental Health Posts using Transformers}


\author{Simranjeet Kaur}

\affiliation{%
  \institution{Thapar Institute of Engineering \& Technology}
  \streetaddress{}
  \city{Patiala}
  \state{Punjab}
  \country{India}
  \postcode{}
}
\email{minhas22june@gmail.com}
\orcid{}

\author{Ritika Bhardwaj}

\affiliation{%
  \institution{Thapar Institute of Engineering \& Technology}
  \streetaddress{}
  \city{Patiala}
  \state{Punjab}
  \country{India}
  \postcode{}
}
\email{rikitabhardwaj600@gmail.com}

\author{Aastha Jain}

\affiliation{%
  \institution{Thapar Institute of Engineering \& Technology}
  \streetaddress{}
  \city{Patiala}
  \state{Punjab}
  \country{India}
  \postcode{}
}
\email{ajain_mca21@thapar.edu}

\author{Muskan Garg}
\affiliation{%
  \institution{University of Florida}
  \streetaddress{}
  \city{Gainesville, FL}
  \country{USA}}
\email{muskanphd@gmail.com}

\author{Chandni Saxena}
\affiliation{%
  \institution{The Chinese University of Hong Kong}
  \city{Hong Kong}
  \country{SAR}
}
\email{csaxena@cse.cuhk.edu.hk}


\begin{abstract}
With recent developments in digitization of clinical psychology, NLP research community has revolutionized the field of mental health detection on social media. Existing research in mental health analysis revolves around the cross-sectional studies to classify users' intent on social media. For in-depth analysis, we investigate existing classifiers to solve the problem of causal categorization which suggests the inefficiency of learning based methods due to limited training samples. To handle this challenge, we use transformer models and  demonstrate the efficacy of a pre-trained transfer learning on "CAMS" dataset~\cite{garg2022cams}. The experimental result improves the accuracy and depicts the importance of identifying cause-and-effect relationships in the underlying text.
\end{abstract}





\maketitle
\section{Introduction}
The NLP research community examines self-reported social media texts to develop AI models highlighting the signs of mental health disorders. In existing literature, mental disorder classification tasks reveal publicly available datasets~\cite{harrigian2021state}. From the previous studies, we comprehend minimal focus on finding the reason behind users' mental state. Consider a post $A$: 
\begin{quote}
    "\textit{With this 2 years of unemployment, I want to quit my life.}"
\end{quote}

In post $A$, a person explicitly mentions about the situation of \textit{'joblessness'}, thereby indicating the causal category (the reason behind their poor mental health) as \textit{'jobs and careers'}.~\citet{son2018causal} attempts to find a solution for causal explanation analysis on social media for small Facebook dataset which is publicly unavailable~\cite{son2018causal}. Therefore, we define the problem of causal categorization on social media posts depicting mental health disorders, and conduct diverse experiments on publicly available dataset: "CAMS"~\cite{garg2022cams}.

\begin{figure}
    \centering
    \includegraphics{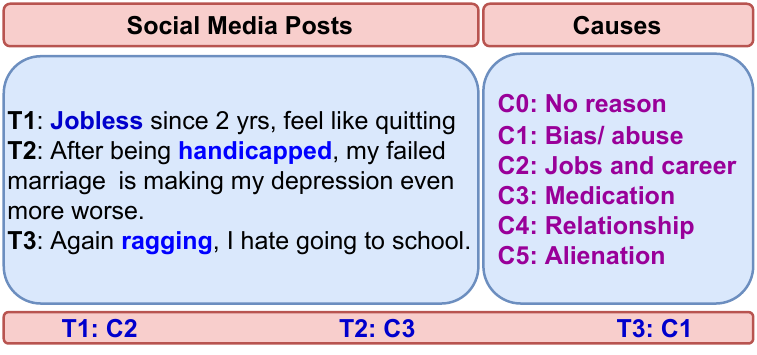}
    \caption{Overview of the architecture: Finding right causal category (C) for any social media post: T1, T2 and T3.}
    \label{fig:1}
\end{figure}

\subsection{Motivation}
As per reports released in August 2021\footnote{https://www.theguardian.com/society/2021/aug/29/strain-on-mental-health-care-leaves-8m-people-without-help-say-nhs-leaders}, \textit{1.6 million people} in England were on waiting lists for mental health care. As per estimation, \textit{8 million people} were not considered \textit{sick enough} to qualify and therefore, they were unable to seek treatment by mental health practitioners. Such obstacles emphasise the importance of automation in identifying users at risk of poor mental health and finding reasons behind it. 

Existing AI models use surface-level linguistic features and semantic-level aspects to classify mental health posts\cite{garg2021survey}. We infer the inability of learning-based models to address the problem of handcrafted features evolved from limited training dataset. To handle this challenge, we use transformer models and test the robustness of transfer learning for multi-class classification. The length of Reddit posts determines the number of words in a post and it varies from few words to thousands of words. We observe that the significant amount of information, intent and cause is expressed in initial part of person writings to raise reader's interest. We hypothesize the sufficiency of the first few tokens for causal categorization and use pre-trained transfer learning with max-length for tokenization as $256$ for our experiments. 

\begin{table*}[t]
    \centering
 
    \begin{tabular}{l|ccc|ccc|ccc}
        \hline
         \textbf{Class} & \multicolumn{3}{c}{\textbf{Crawled corpus}} &
            \multicolumn{3}{c}{\textbf{SDCNL Training data}} &
             \multicolumn{3}{c}{\textbf{SDCNL Test data}}               \\
        
          & \textbf{Min} & \textbf{Max} & \textbf{Avg} & \textbf{Min} & \textbf{Max} & \textbf{Avg} & \textbf{Min} & \textbf{Max} & \textbf{Avg} \\
          \hline
         \textit{No reason} & 1 & 508 & 59.78 & 1 & 1785 & 68.58 & 1 & 1562 & 84.85   \\
         \textit{Bias or Abuse} & 6 & 2109 & 347.48 & 5 & 4378 & 227.24 & 6 & 578 & 149.80  \\
         \textit{Jobs and career} & 13 & 2258 & 228.28 & 17 & 2771 & 255.70 & 20 & 1481 & 206.95   \\
         \textit{Medication} & 5 & 1552 & 213.83 & 3 & 3127 & 205.86 & 11 & 1124 & 165.60  \\
         \textit{Relationship} & 2 & 3877 & 229.35 & 14 & 2739 & 240.08 & 9 & 756 & 202.56 \\
         \textit{Alienation}  & 3 & 1592 & 153.86 & 1 & 899 & 147.01 & 12 & 683 & 145.67 \\
         \hline
    \end{tabular}
    \caption{Word 
    length variation in posts 
    across causal classes
    for each 
    dataset.
    }
    \label{tab4}
\end{table*}

\subsection{Background}
We relate our work on causal categorization with existing research work on causal explanation analysis on social media where~\citet{son2018causal} collect and annotate Facebook data for identifying posts which reflect the cause behind intent of a user. They further extract causal explanations using BiLSTM thereby analyzing (i) the classification: causal posts and non-causal posts, (ii) causal explanations from causal posts. Although, this path-breaking work gives new research directions for causal analysis and finding discourse relations~\cite{son2021discourse}, the major limitations are (i) unavailability of dataset for future research, (ii) limited dataset, (iii) no categorization of causal explanations. To handle this research gap, we perform our analysis on publicly available "CAMS"~\cite{garg2022cams} dataset and solve the problem of causal categorization.

\subsection{Our Contributions}
In association with a senior clinical psychologist, we carry out experiments and evaluate in-depth nuances for causal categorization. The major contributions of this work are
\begin{enumerate}
     \item While deploying learning-based algorithms for causal categorization of mental illness posts, we identify the challenge of limited amount of information for training models.
    \item We hypothesise better efficacy of pre-trained transformers with transfer learning for this task. 
    \item We further investigate prevailing challenges during evaluation and the need for cause-and-effect relationship.

\end{enumerate}
We further categorize our work in different sections. Section 2 explains the materials (dataset) and methods (pre-trained models) required. We explain the results and analyse them in Section 3. We also give possible research directions and reflect the position of research for this task. Finally, we conclude our work in Section 4. 

\section{Materials and Methods}
In this section, we define the problem task and dataset for causal categorization. We further discuss the architecture of the proposed work. As we deploy four different transfer learning models, we brief out the reason behind using each of these one by one. 

\subsection{Problem Definition}
We define the problem of causal categorization as a multi-class classification problem of segregating social media text into one of the five causal categories or into a \textit{no-reason} category. Consider a list of social media post as $L$ where $L= {p_1, p_2, p_3,..., p_n}$ and $p_i$ is the $i^{th}$ post. We consider list of six classes for causal categories as $C = {c_0, c_1, c_2, c_3, c_4, c_5}$ where $c_0$: No reason, $c_1$: bias/ abuse, $c_2$: jobs and careers, $c_3$: medication, $c_4$: relationship, and $c_5$: alienation. In this work, we conduct experiments to build an AI model for classifying $p_i$ into one of the classes in list $C$. 

\subsection{Dataset}
We use annotated "CAMS" dataset having causal categories for each social media post. "CAMS" dataset consists of $5051$ instances classified into either having no-reason or one of the five different causes. Three trained graduate students are deployed to annotate the dataset after professional training given by a senior clinical psychologist. The annotations are verified and the score of Fliess' Kappa inter-observer agreement study is $61.28\%$ for "CAMS" dataset. There are three files in this dataset: the crawled corpus, the SDCNL training corpus and the SDCNL testing corpus. "CAMS" contains annotated SDCNL dataset~\cite{haque2021deep} for causal categorization to test the robustness of the proposed experts guidelines.

In the "CAMS" dataset, the number of instances in every category varies and thus, the dataset is imbalanced. As a result, the existing work on this dataset use training dataset and add $120$ samples for each of the under-represented categories: $c_1, c_2, c_3$ to implement learning-based methods for multi-class classification~\cite{garg2022cams}. The \textit{logistics regression} and \textit{CNN-LSTM} outperforms all other machine-learning and deep-learning mechanisms, respectively. We use them as baselines to propose pre-trained classifiers by hypothesizing the significance of initial part of Reddit post irrespective of length variations as depicted from Table~\ref{tab4}.
\begin{table*}[]
    \centering
    
    \begin{tabular}{cccccccc}
         \hline
         \textbf{Classifier} & \textbf{F1: C0} & \textbf{F1: C1} & \textbf{F1: C2} & \textbf{F1: C3} & \textbf{F1: C4} & \textbf{F1: C5} & \textbf{Accuracy}\\
         \hline
         CNN+LSTM & 0.54 & 0.22 & 0.54 & 0.47 & 0.54 & 0.47 & 0.48\\
         LR & 0.63 & 0.28 & 0.54 & 0.46 & 0.46 & 0.53 & 0.50\\
         DistilBERT & 0.60 & 0.33 & 0.61 & 0.59 & 0.56 & 0.60 & 0.56\\
         BERT-emotion & 0.62 & 0.40 & 0.68 & 0.68 & 0.55 & 0.56 & 0.57\\
         RoBERTa & 0.00 & 0.46 & 0.69 & 0.54 & 0.61 & 0.59 & 0.60\\
         XLNet & 0.69 & 0.47 & 0.72 & 0.59 & 0.68 & 0.64 & \textbf{0.65}\\

         \hline
    \end{tabular}
    
    \caption{Experimental results with "CAMS" dataset. F1 is computed for all six causal classes:  `No reason' (C0), `Bias or abuse' (C1), `Jobs and careers' (C2),  `Medication' (C3), `Relationship' (C4), `Alienation' (C5). 
    }
    \label{tab:results}
\end{table*}
\subsection{Architecture}
The Reddit text is given as an input and after preprocessing, we tokenize the text to perform multi-class classification. We notice diversity in the number of words which varies from a single digit to four digit for all three sets: crawled corpus, SDCNL training data, and SDCNL testing data as shown in Table~\ref{tab4}. However, the consistency in the average number of words prevails for each class in all the three sets as evident from data with an exception of cause 1 ($c_1$) having in-different behaviour thereby showing inconsistency in existing results~\cite{garg2022cams}. This variation reduces the efficiency of deep learning and machine learning mechanisms. We hypothesise that in-depth computations with transfer learning may obtain better insights due to extensive pre-training. Thus, we investigate some special pre-trained models for social media and compare them with baselines.  

\subsection{Pre-trained Models}
We use four pre-trained transformers for this task of multi-class categorization, namely, DistilBERT~\cite{sanh2019distilbert}, BERT-emotion, RoBERTa~\cite{liu2019roberta}, and XLNet~\cite{yang2019xlnet}. We use base models with default parameter settings. We set the maximum length for tokenizer encodings as $256$ by making \textit{truncation} and \textit{padding} true. We further set the Adam optimizer learning rate as $5e-5$ with loss function as \textit{sparse categorical cross entropy} function. We fix the batch size as $16$. 

\citet{sanh2019distilbert} reduces the size of the BERT model by using a method called DistilBERT to pre-train a smaller general-purpose language representation model. Furthermore, EmotionBERT was introduced by extending BERT to map dialogue into causal utterance pairs for emotion prediction tasks that rely significantly on sentence-level context-aware knowledge~\cite{huang2019emotionx}. 

\citet{liu2019roberta} discovered that training the model for longer periods of time, using larger batches of data, deleting the next sentence prediction target, training on longer sequences, and dynamically modifying the masking pattern applied to the training data may all significantly enhance performance. They introduce robustly optimized BERT approach (RoBERTa), which are thus appropriate for long social media texts. \cite{yang2019xlnet} introduce the neural architecture of the XLNet model which is designed to work in association with the auto-regressive goal, including the integration of Transformer-XL and the careful designing of the two-stream attention mechanism.

\section{Experiments and Evaluation}
In this section, we define the baseline models and other pre-trained models to fine-tune them for this task of causal categorization. We further examine our results and deduce inferences for the given task. 

\subsection{Baseline models}
Earlier, the machine learning and deep learning methods were deployed on "CAMS" dataset for multi-class categorization~\cite{garg2022cams}. We use the earlier best performing models: logistic regression and CNN-LSTM as baselines for this task of causal categorization. The logistic regression with softmax function is a solution to multi-class classification. We use default parameters for logistic regression which exhibits best performance as the machine learning approach. A hybrid deep-learning approach, CNN-LSTM, gives the second best results when deployed on "CAMS" dataset. We examine diversity in the behaviour of feature vectorization due to length inconsistency and thus, we exploit transfer learning for long social media texts.

\subsection{Experimental Results}

We demonstrate the experiments for causal categorization using pre-trained transformers for multi-class classification. The results are given in Table~\ref{tab:results} and validate the efficacy of pre-trained transformers over classic machine learning and deep learning mechanisms. We achieve an accuracy of $65\%$ with XLNet and the second best accuracy of $60\%$ with RoBERTa. We further investigate the performance of XLNet, a generalized autoregressive pre-training method. Pretraining with XLNet incorporates \textit{segment recurrence} mechanism and \textit{relative encoding} approach, which empirically increases performance, particularly for jobs involving longer text sequences. XLNet optimizes the estimated log likelihood of a sequence rather than employing a preset forward or backward factorization order as in other auto-regressive models and this bidirectional nature improves causal categorization for long Reddit texts. 

\subsection{Error Analysis}
We obtain significantly improved results for multi-class classification as observed with \textit{students' t-test} at significance level of $5\%$. We compare and contrast the accuracy of each class and observe much variation with the least accuracy of $c_1$ class: bias/ abuse, and maximum accuracy of $c_2$ class: jobs/ careers. This comparative analysis validates a more confused state of \textit{bias and abuse} as compared to the \textit{jobs and careers} class. Our senior clinical psychologists validate this behaviour of classifiers.

We further examine results for each class and notice proportionate improvements for every class except minor deviations in $c_4$ and $c_5$ for DistilBERT. We validate the importance of using auto-regressive models for multi-class classification of long texts. Furthermore, the informal nature of long Reddit texts is a major challenge of natural language processing. 

\subsection{Outlook}
With these observations, we give following future directions to solve the problems of causal categorization:

\begin{itemize}
    \item \textit{Cause-and-effect relationship}: We consider the Reddit posts which determines the users' mental health disorders as intent. The annotators investigate the reason behind their intent to allocate a causal category using underlying guidelines given by experts. To simulate this process, we pose the need of an essential element of cause-and-effect relationship to find discourses.
    \item \textit{Informal social media text}: The user-generated social media data is informal and contain abbreviations, slang and other typographical errors. Embraced with natural language processing, the parsing of lengthy social media posts becomes a challenging task. 
    \item \textit{Explainable AI models}: A potential solution to real-time problem of causal categorization may rely on explainable and deployable multi-class classifiers. We highlight the need of identifying appropriate features for constructing responsible classifiers and encourage this research direction for feasible and transparent solutions. 
    \item \textit{Longformers}: We encourage the use of Longformers\cite{beltagy2020longformer} to analyze the causal categorization of lengthy Reddit posts for cause. However, we recommend the infusion of domain-specific knowledge for experiments and evaluation on "CAMS" dataset.
\end{itemize}

\subsection{Ethical Considerations}
NLP researchers are responsible for transparency about computational research with sensitive data accessed during model design and deployment. We understand the significance of ethical issues while dealing with a delicate subject of mental health analysis. We use the publicly available dataset and do not plan to disclose any sensitive information about the stakeholders (social media users) thereby preserving the privacy of a user~\cite{conway2014ethical}.

We use publicly available pre-trained base models for our demonstration to avoid any ethical conflicts. We assure that we adhere to all ethical guidelines to solve this task. Development of fair AI technologies in mental healthcare supports unbiased clinical decision-making~\cite{uban2021explainability}. Our research work is fair and there is no intentional biasing as we consider the real-time scenario of an imbalanced number of samples for causal categories.

\section{Conclusion}
In this research work, we resolve the problem of causal categorization for social media posts associated with mental health disorders. We fix the issue of training model over limited data in Reddit social media posts by carrying out transfer learning to fine-tune pre-trained AI models. We achieve the accuracy of $65\%$ which is improved by $30\%$ from the previous best performing approach \textit{logistic regression}. We perform error analysis and give possible research directions to improve the transfer learning for real-time use.

\bibliographystyle{ACM-Reference-Format}
\bibliography{reference}

\end{document}